%% file: main.tex
\documentclass{notices}
\usepackage[utf8]{inputenc}
\usepackage{theorem,ifthen,algorithm,algorithmic}
\usepackage{amssymb,amsfonts,amsmath,latexsym,dsfont}
\usepackage{fullpage}
\usepackage{graphicx}
\usepackage{mathrsfs}

\newif\ifAMS
\AMStrue
\AMSfalse

\input{abbrv.tex}

\graphicspath{{images/}{./}}

\title{Differential Equations for Continuous-Time Deep Learning}
\author{Lars Ruthotto\affil{
    Lars Ruthotto is a Winship Distinguished Research Associate Professor at Emory University. His email address is lruthotto@emory.edu.
    }}

\begin{document}
\maketitle
\ifAMS
\else
\begin{abstract}
This short, self-contained article seeks to introduce and survey continuous-time deep learning approaches that are based on neural ordinary differential equations (neural ODEs). 
It primarily targets readers familiar with ordinary and partial differential equations and their analysis who are curious to see their role in machine learning.
Using three examples from machine learning and applied mathematics, we will see how neural ODEs can provide new insights into deep learning and a foundation for more efficient algorithms. 
\end{abstract}
\fi
\section*{}
Deep learning has been behind incredible breakthroughs, such as voice recognition, image classification, and text generation.
While these successes are undeniable, our mathematical understanding of deep learning is still developing.
More rigorous insight is needed to overcome fundamental challenges, including the interpretability, robustness, and bias of deep learning, and lower its environmental and computational costs.

Let us define deep learning informally as machine learning methods that use feed-forward neural networks with many (i.e., more than a handful) layers. 
While most traditional approaches use a finite number of layers, we will focus on more recent approaches that conceptually use infinitely many layers.
We will explain those approaches by defining differential equations whose dynamics are modeled by trainable neural network components and whose time roughly corresponds to the depth of the network.

Using three examples from machine learning and applied mathematics, we will see how continuous-depth neural network architectures, defined by ordinary differential equations (ODEs), can provide new insights into deep learning and a foundation for more efficient algorithms. 
Even though many deep learning approaches used in practice today do not rely on differential equations, I find many opportunities for mathematical research in this area.
As we shall see, phrasing the problem continuously in time enables one to borrow numerical techniques and analysis to gain more insight into deep learning, design new approaches, and crack open the black box of deep learning.
We shall also see how neural ODEs can approximate solutions to high-dimensional, nonlinear optimal control problems.

This article targets readers familiar with ordinary and partial differential equations and their analysis and who are curious to see their role in machine learning.
Therefore, rather than drawing a complete picture of state-of-the-art machine learning, which is shooting at moving targets, we seek to provide foundational insights and motivate further study.
To this end, we will inevitably take shortcuts and sacrifice being up-to-date for clarity.
Even though designing efficient numerical algorithms is critical to translating the theoretical advantages of the continuous-time viewpoint into practical learning approaches, we will keep this topic for another day and provide some references for the interested reader.

While giving a complete picture of the research activity at this interface between differential equations and deep learning is beyond the scope of this paper, we seek to capture the key ideas and provide a head start to those eager to learn more. 
This paper aims to illustrate the foundations and some of the benefits of continuous-time deep learning using a few handpicked examples related to the author's activity in the area.
\ifAMS
The article is written to be self-contained, and it has been an enormous challenge to limit the number of citations to those we believe are most valuable to enable the interested reader an efficient way into this new field.
We provide an unedited arXiv version of this paper to provide additional references and context.
\else
This article is an unedited and extended version of an article under review at AMS Notices and contains a few additional references.
\fi

\section*{Deep Neural Networks in Continuous Time}
Before we begin, let us introduce the main mathematical notation of deep learning for this paper and motivate continuous-time architectures.

Throughout this paper, $\Ftheta: \R^n \to \R^m$ denotes a neural network with the subscript $\theta\in\R^p$ representing its parameters, often called weights. 
We will see that there are different ways of defining $\Ftheta$.
For example, a common choice is the $L$-layer feed-forward network that maps the input  $\x_0 = \x\in\R^n$ to the output $\Ftheta(x) = \x_{L}$ via the layers
\begin{equation}\label{eq:MLP}
    x_{l+1} = \sigma_\ell\left(W_{l} \x_l + b_l \right), \; \forall l = 0, \ldots, L-1.
\end{equation}
Here, the activation functions $\sigma_{l}$ are applied element-wise, and the parameters of the $l$th layer, $\theta_{l}$, consist of the weight matrix $W_{l}$ and the weight vector $b_{l}$.
To define $\Ftheta$, one must choose the number of layers, activation functions, and the sizes of the weight matrices and vectors, collectively called hyperparameters. 
Except for the number of columns in $W_0$, which must be $n$, and the number of rows in $W_{L-1}, b_{L-1}$, which must be $m$, the remaining sizes can be adjusted arbitrarily.

With an effective way to choose hyperparameters and identify the model weights, deep networks are perhaps the most efficient high-dimensional function approximators today.
Their versatility has enabled their use across various tasks.
For example, neural networks have been used in supervised learning to fit given inputs to corresponding outputs, in reinforcement learning to predict optimal actions, and in generative modeling to match simple latent distributions to a complex distribution available only through samples.
The success of neural networks across these tasks is rooted in their approximation properties.
For example, it can be shown that networks with one hidden layer are universal approximators.
Since we will use these networks as an important building block later, let us define them as
\begin{equation}\label{eq:doubleLayer}
    \ftheta(\x) = W_1 \tanh ( W_0 \x + b_0) + b_1, 
\end{equation}
whose parameter, $\theta$, consists of the weight matrices $W_0 \in \R^{k\times n}, W_1\in\R^{m\times k}$, weight vectors $b_0\in\R^k, b_1 \in \R^m$, and whose activation is the hyperbolic tangent function.
In other words, for every $\epsilon>0$, there is a width, $k$, such that they approximate continuous functions to a given accuracy $\epsilon>0$. 
Since the width needed to achieve the desired accuracy can be impractically large, most applications today prefer narrower but deeper architectures.
\ifAMS
\else
This is supported by theoretical results such as in~\cite{KidgerLyons2019}, which show that as $L$ grows, the model becomes more expressive and can be a universal approximator even with a finite width.
\fi

Increasing the depth of neural networks such as the one in~\eqref{eq:MLP} to realize approximation results is easier said than done.
In practice, it often becomes more and more challenging to identify model weights that accurately approximate the function of interest.
For example, it is difficult to approximate the identity function with a network like~\eqref{eq:MLP}.

Residual neural networks (ResNets) provide an alternative way to define very deep networks and considerably improved the state-of-the-art in computer vision applications recently~\cite{He2016identity}.
Their key innovation is often called a skip connection that turns, for example,~\eqref{eq:doubleLayer} into the residual layer
\begin{equation}\label{eq:ResNet}
    \rtheta(\x) 
    % = \x +  W_2 \sigma ( W_1 \x + b_1) + b_2 
    = \x + \ftheta(\x).
\end{equation}
Such a ResNet layer can learn the identity map by the trivial choice $f_\theta \equiv 0$.
Consequently, increasing the network depth by adding residual layers often improves the approximation result since the weights of the new layers can be chosen to approximate the identity.

One way to motivate deep neural networks that are continuous in time is to view $\rtheta(x)$ in~\eqref{eq:ResNet} as a forward Euler approximation of $z(1)$ where $z$ solves the initial value problem
\begin{equation}\label{eq:NODE}
    \frac{d}{dt} z = f_{\theta(t)}(z), \quad t \in (0,1], \quad z(0) = x.
\end{equation}
Here, $t$ is an artificial time, and with the notation $\theta(t)$, we seek to suggest that the weights can be modeled as functions of time; see ~\cites{E:2017kz, HaberRuthotto2017}. 
This viewpoint was popularized in the machine learning community by the work~\cite{Chen:2018vz}, which also coined the term Neural ODEs and demonstrated several new use cases.
It is important to remember that ResNets and Neural ODEs are different: one is discrete, and the other is continuous.
\ifAMS
This has implications both in theory and in practice.
\else
Their relation has been discussed and analyzed in more detail in~\cites{massaroli2020dissecting,Sander2022}.
We also note that when the features, $x$, represent functions (e.g., voice, image, or video data) the above model can mimic partial differential equations~\cite{Ruthotto2018DeepNN}.
\fi

\section*{Supervised Learning in Continuous Time}

In this section, we show how using continuous-time models for supervised learning leads to learning problems that can be analyzed and solved using tools from optimal control.

In supervised learning, the goal is to learn  $\Ftheta$ that approximates the relation between labeled input-output pairs $(\x,\y)\sim D$ assumed to be independent samples from some data distribution $D$. 
Once the hyperparameters of the network are chosen, learning the weights $\theta$ is typically phrased as a minimization problem, such as
\begin{equation}\label{eq:optProb}
    \min_\theta \mathbb{E}_{(x,y)\sim D} [ \ell(\Ftheta(x), y) ],
\end{equation}
with some loss function $\ell$ whose definition depends on the task; for example, for data fitting, one can consider the regression loss $\ell(v,y)=\frac{1}{2} \|v-y\|^2$.
\ifAMS
The optimization problem is challenging and interesting in its own right.
\else
The optimization problem is challenging and interesting in its own right, and we refer to~\cite{bottou2016optimization} for an excellent monograph.
\fi 

A simple way to define a continuous-time model is to define  $\Ftheta$ as an affine transformation of the terminal state of a neural ODE, that is,
\begin{align}\label{eq:arch}
    \Ftheta(x) &= W z(1) + b \\
    \frac{d}{dt} z & = f_{\theta(t)}(z),\quad t \in (0,1], \quad 
    z(0) = x.
\end{align}
This model can be interpreted as the affine model (given by the parameters $W$ and $b$) applied to the features evolved by the neural ODE, whose dynamics are governed by the weight function $\theta$.
Note that the affine function can be omitted when $m=n$.

\begin{figure*}[ht]
    \centering
    \includegraphics[width=0.9\textwidth]{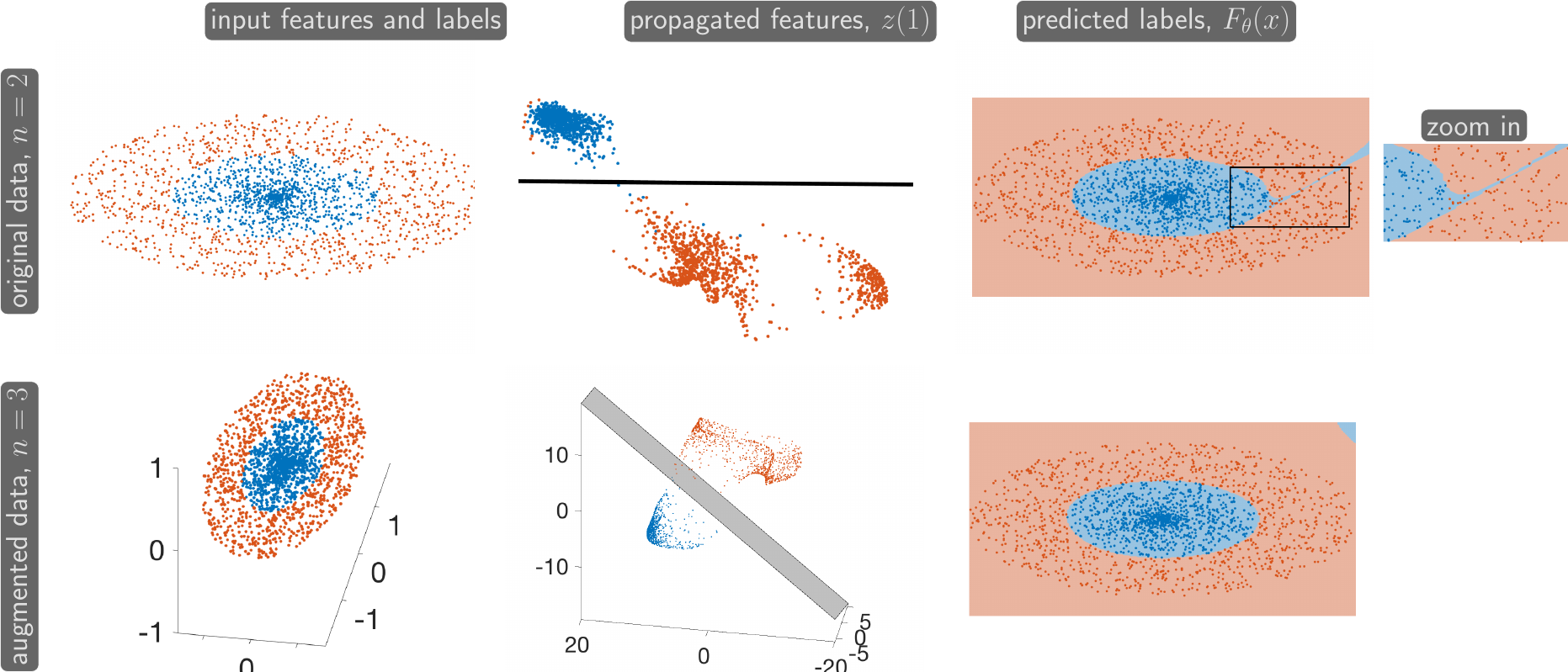}
    \caption{Illustration of a continuous-time model for binary classification. Left column: input features and labels. Center column: propagated features given by the final state of ODE and hyperplane given by $W,b$. Right column: Labels predicted by the neural network. The rows show two instances of the problem using the original two-dimensional and augmented features (padded with one zero), respectively. While both models agree closely around the samples, we highlight some errors of the original model that arise from the restriction to invertible maps in $n=2$. This example demonstrates that augmenting overcomes the need for a non-invertible mapping. Note that the models may not be reliable in regions with no data points, for example, in the top right corner of the domain. \label{fig:NODE} }    
\end{figure*}

Since neural ODEs as in~\ref{eq:arch} yield an invertible transformation of the data space, it is unsurprising that many functions cannot be approximated by the model in~\eqref{eq:arch}.
In Figure~\ref{fig:NODE}, we demonstrate this in one example that can also illustrate the role of the ODE in the supervised learning problem.
Shown here is a classification example obtained by minimizing a logistic regression loss function.
Each data point $\x$ is associated with a label $\y$, either blue or red.
The goal is to learn a function $\Ftheta$ corresponding to the training data.
The center column of the figure shows the propagated features, which are the $z(1)$ associated with each example, as well as the hyperplane parameterized by $W,b$ that seeks to divide the features. 
The rightmost column visualizes the predictions of the classifier. 
While the predictions are nearly perfect in both rows, upon close inspection, the limitations of the ODE shine through in the top row, and it can be seen that the network was unable to transform the blue and red points to become linearly separable, which requires a non-invertible transformation. 
The need for that is alleviated by simply padding the input features with one zero and embedding them into three dimensions. 
This phenomenon and the importance of augmentation are elaborated in~\cite{Dupont:2019wa}.

When  $\Ftheta$ is defined as in~\eqref{eq:arch} the minimization problem~\eqref{eq:optProb} becomes an optimal control problem
\begin{equation}\label{eq:NODEasOC}
\begin{split}
    \min_{\theta,W,b,z} \; & \mathbb{E}_{(x,y)\sim D} [ \ell(W z(1) + b, y) ],\\
    \subto \; & \frac{d}{dt} z  = f_{\theta(t)}(z),\quad t \in (0,1]\\   & z(0) = x.        
\end{split}
\end{equation}
The relation between learning $\theta$ and solving an optimal control problem has been used to gain insights and obtain more efficient algorithms. 
Universal approximation results of continuous-time models are derived in~\cite{Li:2019wr} by analyzing their flow maps. 
On the computational side, it is worth mentioning the approach in ~\cite{Li:2017wr}, which uses control theory to derive new learning algorithms, and~\cite{benning2019deep}, which studies the continuous and discrete versions of the learning problem and proposes schemes that can learn time discretizations and embed other constraints. 

With an architecture continuous in time, it is also possible to provide a PDE perspective of the supervised learning problem~\eqref{eq:NODEasOC}.
Following the presentation in~ % \cite{li2018deep,WangEtAl2018}
\cite{WangEtAl2018}, let us take a  macroscopic viewpoint and consider the space of examples rather than individual data points.
To this end, we model the neural network predictions as a function $u : \R^n \times [0,1] \to \R^m$ whose evolution is governed by the transport PDE with velocity $\ftheta$.
In doing so, we formulate supervised learning as a PDE-constrained optimization problem
\begin{equation} 
\begin{split}\label{eq:NODEasPDECO}
    \min_{u,\theta,W,b}  \;&\; \mathbb{E}_{(x,y)\sim D} [\ell(u(x, 1), y) ], \\
    \subto\; &\; \partial_t u(z,t) + f_{\theta(t)}(z)^\top \nabla u(z,t)  = 0  \\
    \;& \; u(z,0)   = W x + b.    
\end{split}
\end{equation}
To verify that the problems are equivalent, note that~\eqref{eq:arch} defines the characteristic curves of the transport equation and therefore 
\begin{align*}
    u(x, 1) & = u(z(0),1)  \\
            & = W z(1) + b  \\
            &  = \Ftheta(x) \approx y.
\end{align*}
Since the dimensionality of the feature space is usually larger than two or three, problem~\eqref{eq:NODEasPDECO} is intractable. However, this viewpoint can provide new insights and motivate improved learning algorithms. 
For example,~\cite{WangEtAl2018} showed that adding some amount of viscosity to the PDE constraint in~\eqref{eq:NODEasPDECO} can increase the robustness of classifiers to random perturbations of the inputs. 
They also proposed an efficient method based on the Feyman-Kac formula to scale to high dimensions.
The interpretation also allows us to build bridges to optical flow and image registration, which may yield further understanding in the future.

\section*{Continuous-Time Generative Models}

This section illustrates the advantages of continuous-time models for building flexible generative models.

In generative modeling, one learns complicated, often high-dimensional, data distributions from examples.
The goal typically is to enable sampling and sometimes includes estimating the densities of given examples.
Rather than mapping input points to corresponding output points, as in supervised learning, a generative model maps a tractable reference distribution to a target distribution.
In deep generative modeling, this mapping is called a generator, and it is represented by a deep neural network.

\begin{figure}
    \centering
    \includegraphics[width=0.4\textwidth]{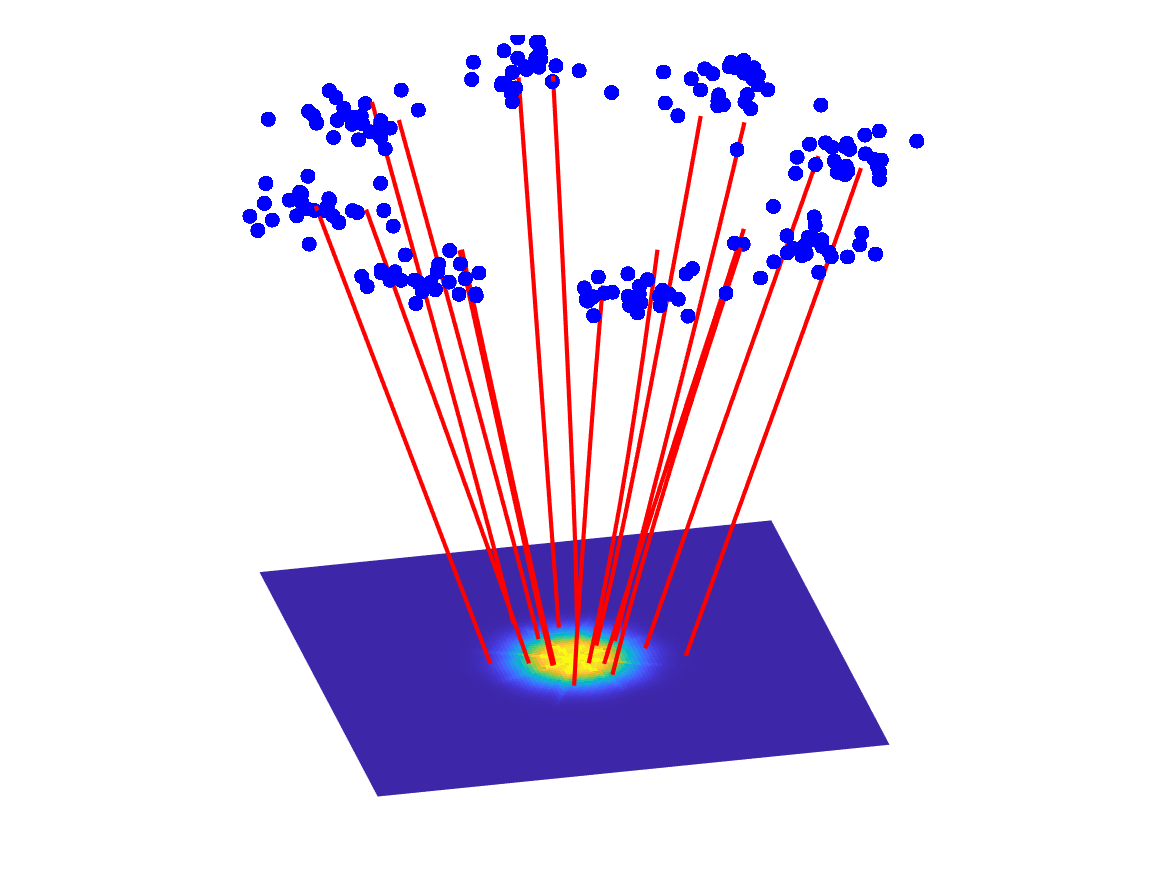}
    \caption{Illustrating the generative modeling problem. Given samples from the target distribution (represented by blue dots), we try to find an invertible transformation (represented by red lines) to a simple target distribution (a standard Gaussian).}
    \label{fig:CNFoverview}
\end{figure}

While the choice of reference distribution is arbitrary (as long as it is easy to sample from), most commonly, it is a standard Gaussian.
To motivate the use of continuous-time models, we will set the dimension of the latent distribution, $n$, equal to the data distribution, $m$.
We also assume both distributions are proper in $\R^n$, which enables the use of normalizing flows
\ifAMS
.
\else
; see \cite{Kobyzev2020NormalizingFA} for a recent review.
\fi
When this assumption is violated (as is common in practice), other generative models, such as variational autoencoders or generative adversarial networks, are typically superior; a general introduction to generative modeling is given in~\cite{RuthottoHaber2021}.
We illustrate the generative modeling problem in Figure~\ref{fig:CNFoverview}.

Under these assumptions, the idea of normalizing flows is to learn a diffeomorphic generator that maps reference to target; that is, we seek to find an invertible $\Ftheta$ such that both $\Ftheta$ and $\Ftheta^{-1}$ are continuously differentiable. 
Since the reference distribution is a standard Gaussian, its density, which we denote by $\pi_X$, is easy to compute.
To estimate the density of a  point $y$ from the unknown target distribution under the current generator $\Ftheta$, we can use the change of variable formula
\begin{equation}
    \pi_Y(y) = \pi_X(\Ftheta^{-1}(y)) \cdot \det\nabla \Ftheta^{-1}(y).
\end{equation}
Maximum likelihood training aims to find a parameter $\theta$ that maximizes the expected likelihood over all the samples.
One typically considers minimizing the expected negative log-likelihood 
\begin{equation}\label{eq:MLE}
     \mathbb{E}_{y} \left[ \hf \|\Ftheta^{-1} (y)\|^2 - \log\det\left(\nabla \Ftheta^{-1}(y)\right)\right],
\end{equation}
where the first term is (up to an additive constant) the negative log-likelihood of the standard normal $\pi_X$.
Even though the functional is convex in $\Ftheta^{-1}$, it is not convex in $\theta$ once we approximate it with a neural network.
Therefore, numerical optimization schemes are required to compute an approximate minimizer.

The most critical question for normalizing flows is choosing the neural network architecture $\Ftheta$.
For example, even though the classical multi-layer perception~\eqref{eq:MLP} can approximate any generator, it is generally not invertible, so it cannot be trained using~\eqref{eq:MLE}.
Trading off the expressiveness of the model with invertibility and computational considerations has led to various approaches. 
A key idea to build normalizing flows is to concatenate finitely many layers designed to have easy-to-compute inverses and Jacobian log determinants.
However, this construction can limit expressiveness, and often, many layers are needed to approximate mappings in high dimensions. 
As discussed in more detail in~\cite{RuthottoHaber2021}, one can sometimes increase the approximation power of $\Ftheta^{-1}$ by sacrificing the computational efficiency of evaluating the generator~$\Ftheta$ or vice versa.

Approaches that define the generator using a continuous-time model are known under the term continuous normalizing flows~\cite{GrathwohlEtAl2018}.
When defining $\Ftheta(x) = z(1)$ as the terminal state of the neural ODE~\eqref{eq:NODE} and $\ftheta$ is sufficiently regular so that the ODE solution is defined uniquely, we can, at least formally, compute the inverse of the generator by integrating backward in time and defining $\Ftheta^{-1}(y) = w(0)$ where
\begin{equation}\label{eq:CNFbwd}
    \frac{d}{dt} w = f_{\theta(t)}(w), \quad t \in [0,1), \quad w(1) = y.
\end{equation}
One should always be careful when reversing the time in an ODE since, in general, it should not be assumed that the ODE is stable both forward and backward. 
However, in practice, using fairly arbitrary neural networks to represent $\ftheta$ has been found to induce negligible errors.

Let us also comment on the evaluation of the Jacobian log-determinant. 
When we define $\Ftheta^{-1}(y) = w(0)$  as above, the instantaneous change of variables formula from~\cite{Chen:2018vz}*{Appendix A}  implies 
\begin{equation}
    \log\det(\nabla \Ftheta^{-1}(y)) = \int_0^1 {\rm tr} \nabla f_{\theta(t)}(w) dt.
\end{equation}
Numerically integrating the trace of the Jacobian using quadrature rules is often computationally more efficient than computing its log determinant.
The computation of the log determinant can also be combined with the numerical ODE solver used to compute the inverse of the generator.

The training of the continuous-time generator $\Ftheta$ can now be phrased as an optimal control problem
\begin{equation}\label{eq:CNF}
\begin{split}
    \min_{\theta, w} \; \mathbb{E}_{y} & \left[ \hf \|w(0)\|^2 - \int_{0}^1 {\rm tr} \nabla f_{\theta(t)}(w)dt \right] \\
    \subto \;&   \frac{d}{dt}  w
    = 
    f_{\theta(t)}(w), \; t \in [0,1),\;\\
    & w(1)=y.
\end{split}
\end{equation}
It turns out that this control problem admits infinitely many solutions.
As we shall see in the following section, it is possible for two different networks $\ftheta^{(1)}$ and $\ftheta^{(2)}$ to yield the same generator but follow different trajectories; impatient readers may skip ahead to Figure~\ref{fig:MFG}.
Even though one often cannot see the differences in the created samples, realizing the non-uniqueness allows one to bias the search toward generators with more regular trajectories.
It also bridges generative modeling and optimal transport, which has a rich theory and long history.

The idea of penalizing transport costs has been investigated and shown practical benefits in~\cites{Yang:2019tj,Finlay:2020wt, OnkenEtAl2020OTFlow}.
Since optimal transport in high dimensions is difficult, a tractable approach is to penalize  transport costs, for example, by adding the functional
\begin{equation}\label{eq:POT}
P_{\rm OT}[w,\ftheta] =   \int_0^1 \frac{\alpha}{2} \|f_{\theta(t)}(w)\|^2 dt
\end{equation}
to the objective function in ~\eqref{eq:CNF}.
Here, the parameter $\alpha$ balances matching the distributions (for $\alpha \ll 1$) and minimizing the transport costs (for $\alpha \gg 0$).
It is possible to show that we obtain the optimal transport map for an appropriate choice of $\alpha$ and that trajectories become straight.
This regularity can translate to practical benefits since the ODEs for the generator and its inverse become trivial to solve. 

Adding transport costs to the generative modeling problem also provides exciting opportunities to analyze the training problem.
To give a glimpse into this area, we note that the training problem can be written on the macroscopic level as the PDE-constrained optimization problem
\begin{equation}\label{eq:dynamicTrainingProblem}
	\begin{split}
		\min_{\rho,\theta}   \int_{\R^n} &\int_0^1\frac{\alpha}{2} \|f_{\theta(t)}(\x)\|^2 \rho(t,\x) dt  - \log \rho(1,\x) \pi_Y(\x)   dx \\
		\subto \;&\;
		\partial_t \rho(t,\x) + \nabla\cdot\left(f_{\theta(t)}(\x) \rho(t,\x)\right)=0, \\ 
		& \quad \rho(0,\x) = \pi_X(\x).
	\end{split}    
\end{equation}
Here, the PDE constraint for $t\in(0,1]$ is given by the continuity equation.
\ifAMS
The formulation above is a relaxed version of the classical dynamic optimal transport formulation.
\else
The formulation above is a relaxed version of the classical dynamic optimal transport formulation~\cite{BenamouBrenier2000}.
\fi
To be concrete, above the terminal constraint $\rho(1,x)=\pi_Y$ is relaxed, and deviations are penalized by the second term of the objective.
Similar to the dynamic OT problem, ~\eqref{eq:dynamicTrainingProblem} can be reformulated into a convex variational problem, which can provide further insight but becomes impractical when the growth of $n$ requires nonlinear function approximators.

\section*{Neural ODEs for Potential Mean Field Games}
This section showcases Neural ODE's promise for overcoming the limitations of other numerical methods for simulating interactions within large populations of agents playing a non-cooperative game.

\begin{figure}[t]
    \centering
    \includegraphics[width=0.45\textwidth]{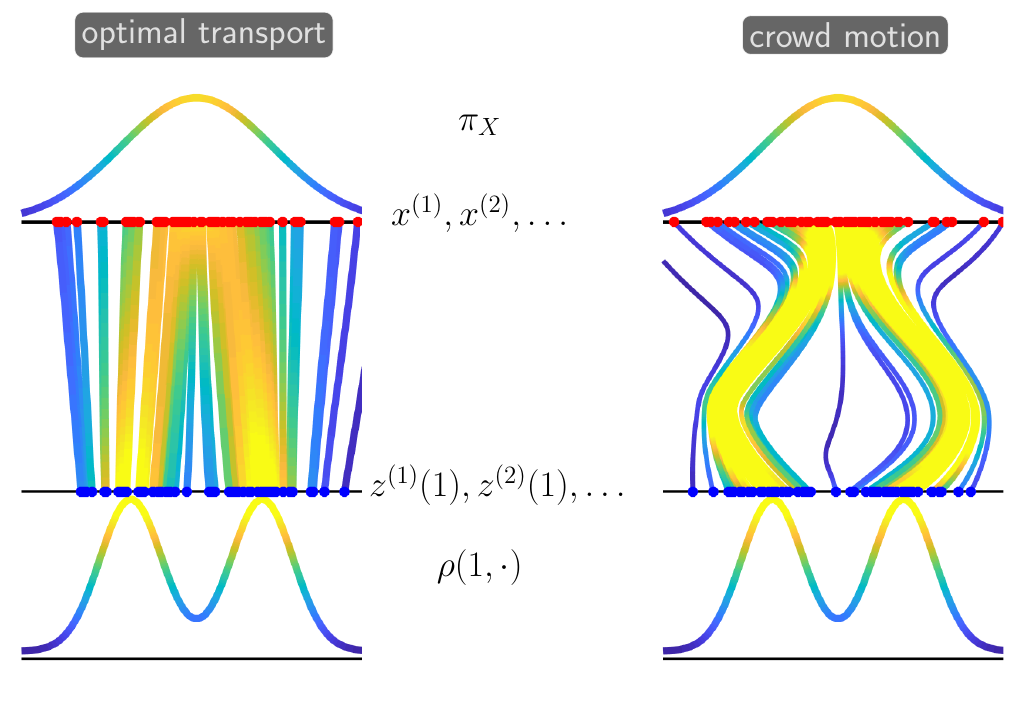}
    \caption{Illustration of potential mean field game versions of relaxed dynamic optimal transport (left) and crowd motion problems (right). Both cases use the standard Gaussian reference $\pi_X$ (top) and the same Gaussian mixture as the target (bottom). As expected, in the optimal transport case, the trajectories are straight, whereas in the crowd motion case, the agents are curved to avoid an obstacle in the center of the domain. This example also shows that different dynamics can produce the same map $\Ftheta$.}
    \label{fig:MFG}
\end{figure}

To show how neural ODEs arise naturally in the mean field limit of many games, let us generalize  \eqref{eq:dynamicTrainingProblem} to include objective functions that contain more general cost terms in, for example, using the objective functional
\begin{equation}
\begin{split}
    \calJ[\rho,\ftheta] = &  \int_{0}^1 \int_{\R^n} L(x,\ftheta) \rho(t,x) dx dt  \\
     & + \int_{0}^1 \calF(\rho(t,\cdot)) dt + \calG(\rho(1,\cdot)),
\end{split}
\end{equation}
which consists of running cost  given by the function $L:\R^n \times \R^n \to \R$ and the functional $\calF$, and a terminal cost functional $\calG$.
Given an initial state of the population density $\pi_X$, finding the optimal strategy $\ftheta$ amounts to solving the mean field game
\begin{equation}\label{eq:MFGmacro}
    \begin{split}
        \min_{\rho,\theta} \;&  \calJ[\rho,\ftheta]\\
        \text{ s.t. }\;& \partial_t \rho(t,\x) + \nabla\cdot\left(f_{\theta(t)}(\x) \rho(t,\x)\right)=0, \\ 
		& \quad \rho(0,\x) = \pi_X(\x),
    \end{split}
\end{equation}
where the continuity equation models the evolution of the population density $\rho$.
This more general version of~\eqref{eq:dynamicTrainingProblem} can be used to model various non-cooperative differential games played by a large population of rational agents.
Furthermore, the formulation allows one to analyze and solve a larger set of generative models beyond continuous normalizing flows~\cite{ZhangEtAl2023}.
Some examples are listed in Table~\ref{tab:MFG}. % List OT, Pedestrian, CNF, score-based, ...
Also, we provide two one-dimensional instances motivated by optimal transport and crowd motion to illustrate our problem setup and notation in Figure~\ref{fig:MFG}.
For simplicity, we focus the observation on deterministic games but note that neural network techniques have also been proposed for stochastic MFGs governed by the Fokker Planck equation~\cite{Lin:2020wv}.
\begin{table*}[t]
    \centering
    \begin{tabular}{|c|c|c|c|}
         & $L(x,\ftheta)$ & $\mathcal{F}(\rho)$ & $\mathcal{G}(\rho)$  \\ \hline
      optimal transport (OT) & $\displaystyle \frac{\alpha}{2}\|\ftheta\|^2$ &  & $\displaystyle{\rm KL}(\rho, \rho_Y)$\\
      crowd motion & $\displaystyle \frac{\alpha}{2}\|\ftheta\|^2 + Q(x)$ & $\displaystyle \int \rho\log(\rho) dx $ & \\
      normalizing flow & & & $\displaystyle \mathbb{E}_y[-\log(\rho)]$  \\
      normalizing flow+OT  & $\displaystyle \frac{\alpha}{2}\|\ftheta\|^2$ & & $\displaystyle \mathbb{E}_y[-\log(\rho)]$  \\ 
    \end{tabular}
    \caption{Examples of different potential mean field games that can be modeled via~\eqref{eq:MFGmacro}. We also recommend~\cite{ZhangEtAl2023} for a more exhaustive list, including score-based diffusion and Wasserstein gradient flows.}
    \label{tab:MFG}
\end{table*}

It is important to note that solving~\eqref{eq:MFGmacro} in general remains a daunting task, especially when the dimension of the state space, $n$, is larger than three or four.
In these cases,  the curse of dimensionality affects traditional numerical methods that rely on meshes or grids to solve the continuity equation in~\eqref{eq:MFGmacro}.
Unfortunately, many realistic use cases of MFGs arising in economics, social science, and other fields require considerably larger $n$ to capture the state of the agents.

Deriving a neural ODE formulation for approximating the solution of a class of  MFGs requires some calculus and theoretical tools.
Here, we will briefly overview our approach in~\cite{RuthottoEtAl2020MFG}.
A key quantity for analyzing and solving the mean field game is its value function $\Phi : \R\times \R^n \to \R$.
An intuitive way to define it is via a microscopic perspective.
Let  $x\in\R^n$ be the state of an arbitrary agent at time $t \in [0,1)$. 
Then, the value function $\Phi$ denotes the optimal cost to go for this agent and can be written as
\begin{equation}\label{eq:MFGmicro}
\begin{split}
    \Phi(t,x) =&  \min_{\theta,z} J_t[\ftheta,\rho,z], \\
    \subto\;   & \frac{d}{dt} z = f_{\theta(s)}(z),\; s \in (t,1]\\
    & z(t)=x
\end{split}
\end{equation}
where $\rho$ is the population density at the equilibrium, and we define the single-agent objective functional 
\begin{equation}\label{eq:JMFGmicro}
\begin{split}
     J_t[\ftheta&,\rho,z] =   G(z(1),\rho(1,z(1)))      \\
     &+  \int_{t}^T \left(L\left(z,f_{\theta(s)}(z)\right) + F(z,\rho(s,z)) \right)ds
\end{split}
\end{equation}
with $F$ and $G$ denoting the $L^2$ derivatives of $\calF$ and $\calG$, respectively.
This makes this an MFG in potential form and ensures that solving the problem from the microscopic and macroscopic perspective leads to the same solution. 

Evaluating $J_t$ in ~\eqref{eq:JMFGmicro} requires the agent to estimate the equilibrium density resulting from the collective behavior of the agents around its current trajectory.
To this end, we solve the continuity equation in~\eqref{eq:MFGmacro}.
Fortunately, characteristic curves of the continuity equation coincide with the trajectories of the agents. 
Hence, rather than solving the continuity equation everywhere to compute the density, we can update the densities along the trajectories as the agent's state evolves. 
Since most of the common choices listed in Table~\ref{tab:MFG} require the log of the density, let us note that along the curves $z(\cdot)$ we have
\begin{equation}
   \log \rho(t,z(t)) =  \log\pi_X(x) - \int_0^t {\rm tr}\nabla f_{\theta(s)}(z)   ds.
\end{equation}
This allows us to eliminate the continuity equation in problem~\eqref{eq:MFGmicro} without requiring a  grid or a mesh.
The resulting Lagrangian approach also has a crucial computational advantage since we can compute the trajectories and objective function values for several agents in parallel without the need to communicate.

The above observations and our experience from the previous section could also be used to obtain a neural ODE approach for potential mean field games.
However, the learned solution may violate some theoretical properties. For example, the Pontryagin Maximum Principle shows that the optimal control, $f^*$, is related to the value function via the feedback form
\begin{equation}\label{eq:feedback}
    f^*(x) = - \nabla_p H(x,\nabla \Phi(t,x)).
\end{equation}
Here, the Hamiltonian $H$ is the Fenchel dual of the running cost $L$  defined by
\begin{equation}
    H(x,p) = \sup_{f \in \R^n} \left\{-p^\top f - L(x,f) \right\}.
\end{equation}
For many choices of $L$ that arise in practice, including the examples in Table~\ref{tab:MFG}, $H$ can be computed analytically.

Even when the neural network can approximate the optimal policy, in our experience, finding weights such that properties such as~\eqref{eq:feedback} approximately hold is non-trivial.
Clearly, choosing the network weights randomly will not yield an approximation that satisfies the feedback form, which means we must solve the learning problem well.  
Therefore, when $H$ is available and straightforward to compute, we can approximate $\Phi_{\theta}$ with a scalar-valued neural network and define the control implicitly via~\eqref{eq:feedback}.
Since the value function contains all the information about the MFG solution, approximating it directly can provide helpful insight.

Approximating the value function directly also allows us to incorporate more prior knowledge into the training problem.
It is known that the value function solves the Hamilton Jacobi Bellman (HJB) equations
\begin{equation}\label{eq:HJB}
\begin{split}
    -\partial_t \Phi_{\theta}(t,x) + H(x,\nabla \Phi_\theta(t,x)) & = F(x,\rho(t,x)),\\
    \Phi_\theta(1,x) &  = G(x,\rho(1,x)).
\end{split}
\end{equation}
Here, the $-\partial_t$ emphasizes that this equation is backward in time. 
\ifAMS
Lasry and Lions have shown that solving the above PDE in conjunction with the continuity equations is the necessary and sufficient condition for the problem.
\else
Lasry and Lions have shown that solving the above PDE in conjunction with the continuity equations is the necessary and sufficient condition for the problem~\cite{LasryLions2007}.
\fi
Even though solving~\eqref{eq:HJB} in high dimensions is affected by the curse of dimensionality, we can monitor the violations of the HJB equations along the trajectories and penalize them using some functional $P_{\rm HJB}$.

To summarize the above observations, we can train the scalar deep neural network, $\Phi_\theta$, that approximates the  value functions via the  optimal control problem
\begin{equation}
    \begin{split}
    \min_{\theta} &\; \mathbb{E}_{x \sim \pi_X} \left[ J\left[\ftheta,\rho,z  \right] + \beta P_{\rm HJB}[\Phi_\theta,\rho,z] \right] \\
    \subto   \;& \;\frac{d}{dt}
    \left(
    \begin{array}{@{}c@{}}
    z\\
    \log(\rho(t,z))
    \end{array}
    \right)
    = 
    \left(
    \begin{array}{@{}c@{}}
    f_{\theta}(z)\\
    -{\rm tr}\nabla f_{\theta}(z)
    \end{array}
    \right)
     \\
    &z(0) = x, \quad \log\rho(0,x)=\log \pi_X(x).    
    \end{split}
\end{equation}

\section*{Discussion and Outlook}

We demonstrated how differential equations can be used to build continuous-time deep learning approaches. We highlighted a few ways this could lead to new insights and more efficient algorithms for supervised machine learning, generative modeling, and solving high-dimensional mean field games.

The key step is to transform the feature space incrementally using an ODE whose dynamics are represented by a deep neural network.
Conceptually, this leads to infinitely deep networks whose artificial time loosely corresponds to the depth of the network.
This relation is not precise since, depending on the choice of $\ftheta$, the dynamics of the ODE can be scaled arbitrarily and even depend on trainable parameters.
One can consider continuous-time architectures as infinitely deep, which sets them apart from traditional networks consisting of finitely many layers.

Earlier works on continuous-time learning that bring in ideas from differential equations include, for example,~\cite{GonzalezGarcia:2004wp}.
Some of the ideas in this work resemble the ones popularized by~\cite{Chen:2018vz}, but the latter contained other novel ideas, for example, the use of differential equations for generative models later extended in~\cite{GrathwohlEtAl2018}.
The renewed interest in continuous-time models is probably related to the growth of computational resources, advances in numerical methods for solving ODEs and optimal control problems arising in their training, and larger datasets.

We inevitably omitted many important topics to keep the presentation short and coherent. 
Important examples of continuous-time architectures not governed by ODEs are the controlled differential equations~\cite{kidger2020neural} and non-local networks driven by fractional differential equations~\cite{antil2020fractional}.
The above viewpoint can be extended to PDE architectures for input features that can be seen as grid functions.

More could also be said about numerical methods for continuous-time deep learning.
An important question is whether first to optimize and then discretize (as, for example, in~\cites{Chen:2018vz, GrathwohlEtAl2018}) or first to discretize and then optimize (as, for example, done in~\cite{OnkenEtAl2020OTFlow}).
In a first-optimize-then-discretize setting, one solves the adjoint ODE to compute the gradient of  the loss function with respect to the weights, the key ingredient for optimization algorithms. 
While this allows some flexibility in choosing the numerical integrators for the forward and adjoint equation (e.g., one can use different step sizes),  the adjoint method requires storing or recomputing the entire trajectory of the features, which is computationally infeasible.
In a discretize-then-optimize approach,  one selects a numerical time integrator and integration points to discretize the neural ODE~\eqref{eq:NODE} and obtains a finite-dimensional optimization problem. 
Differentiating the discretized loss function with respect to the weights is possible using automatic differentiation (also known as back-propagation) or analytically using the chain rule. 
The choice of the time integrator is crucial and provides opportunities to design novel network architectures that resemble ResNets but can be tailored to the network model;  for example, one can mimic hyperbolic systems and use symplectic time integrators to ensure forward and backward stability~\cite{HaberRuthotto2017} and save memory costs.
\ifAMS
\else
An excellent in-depth discussion of these two paradigms in the context of neural ODEs is provided in~\cite{gholami2019anode}. Some additional numerical results for time-series regression and generative modeling can be found in~\cite{onken2020discretize}.
\fi

\section*{Acknowledgments}
I thank $\{$Levon Nurbekyan, Deepanshu Verma$\}$ for their careful reading of initial drafts and fruitful discussions.
I am also grateful for the two anonymous reviewers who made excellent suggestions that improved this paper.
My work on this article and related projects was partly supported by NSF awards DMS 1751636, DMS 2038118, AFOSR grant FA9550-20-1-0372, and US DOE Office of Advanced Scientific Computing Research Field Work Proposal 20-023231.

\bibliographystyle{abbrv}
\bibliography{main}
\end{document}

%% file: abbrv.tex
\newcommand{\hf}{{\frac 12}}

\newcommand{\calF}{{\cal F}}
\newcommand{\calJ}{{\cal J}}
\newcommand{\calG}{{\cal G}}

\newcommand{\R}{\ensuremath{\mathds{R}}}

\newcommand{\x}{x}
\newcommand{\y}{y}
\newcommand{\Ftheta}{F_\theta}
\newcommand{\ftheta}{f_\theta}
\newcommand{\rtheta}{r_\theta}

\newcommand{\subto}{\text{s.t.}}

%% file: main.bbl
\begin{bibdiv}
\begin{biblist}

\bib{antil2020fractional}{article}{
      author={Antil, Harbir},
      author={Khatri, Ratna},
      author={L{\"o}hner, Rainald},
      author={Verma, Deepanshu},
       title={Fractional deep neural network via constrained optimization},
        date={2020},
     journal={Machine Learning: Science and Technology},
      volume={2},
      number={1},
       pages={015003},
}

\bib{BenamouBrenier2000}{article}{
      author={Benamou, J~D},
      author={Brenier, Y},
       title={{A computational fluid mechanics solution to the Monge-Kantorovich mass transfer problem}},
        date={2000},
     journal={Numerische Mathematik},
}

\bib{benning2019deep}{article}{
      author={Benning, Martin},
      author={Celledoni, Elena},
      author={Ehrhardt, Matthias~J},
      author={Owren, Brynjulf},
      author={Sch{\"o}nlieb, Carola-Bibiane},
       title={Deep learning as optimal control problems: Models and numerical methods},
        date={2019},
     journal={Journal of Computational Dynamics},
      volume={6},
      number={2},
       pages={171\ndash 198},
}

\bib{bottou2016optimization}{article}{
      author={Bottou, Léon},
      author={Curtis, Frank~E},
      author={Nocedal, Jorge},
       title={{Optimization Methods for Large-Scale Machine Learning}},
        date={2016},
     journal={SIAM Review},
      volume={60},
      number={2},
       pages={223 311},
      eprint={1606.04838},
}

\bib{Chen:2018vz}{inproceedings}{
      author={Chen, Ricky~TQ},
      author={Rubanova, Yulia},
      author={Bettencourt, Jesse},
      author={Duvenaud, David~K},
       title={Neural ordinary differential equations},
        date={2018},
   booktitle={Advances in neural information processing systems},
      volume={31},
}

\bib{Dupont:2019wa}{inproceedings}{
      author={Dupont, Emilien},
      author={Doucet, Arnaud},
      author={Teh, Yee~Whye},
       title={Augmented neural {ODE}s},
        date={2019},
   booktitle={Advances in neural information processing systems},
      volume={32},
}

\bib{E:2017kz}{article}{
      author={E, Weinan},
       title={{A Proposal on Machine Learning via Dynamical Systems}},
        date={201703},
        ISSN={2194-6701},
     journal={Communications in Mathematics and Statistics},
      volume={5},
      number={1},
       pages={1 11},
         url={message:\%3CDM5PR03MB2843C6F450F4E51E4CC11AA3A0940@DM5PR03MB2843.namprd03.prod.outlook.com\%3E},
}

\bib{Finlay:2020wt}{incollection}{
      author={Finlay, Chris},
      author={Jacobsen, Joern-Henrik},
      author={Nurbekyan, Levon},
      author={Oberman, Adam},
       title={{How to Train Your Neural ODE: the World of Jacobian and Kinetic Regularization}},
        date={202011},
   booktitle={International conference on machine learning},
   publisher={PMLR},
       pages={3154 \ndash  3164},
}

\bib{GrathwohlEtAl2018}{inproceedings}{
      author={Grathwohl, Will},
      author={Chen, Ricky~TQ},
      author={Bettencourt, Jesse},
      author={Sutskever, Ilya},
      author={Duvenaud, David},
       title={{FFJORD}: Free-form continuous dynamics for scalable reversible generative models},
        date={2018},
   booktitle={International conference on learning representations},
}

\bib{gholami2019anode}{article}{
      author={Gholami, Amir},
      author={Keutzer, Kurt},
      author={Biros, George},
       title={Anode: Unconditionally accurate memory-efficient gradients for neural odes},
        date={2019},
     journal={arXiv preprint arXiv:1902.10298},
}

\bib{HaberRuthotto2017}{article}{
      author={Haber, Eldad},
      author={Ruthotto, Lars},
       title={{Stable architectures for deep neural networks}},
        date={2017},
        ISSN={0266-5611},
     journal={Inverse Problems},
      volume={34},
      number={1},
       pages={014004},
      eprint={1705.03341},
}

\bib{He2016identity}{inproceedings}{
      author={He, Kaiming},
      author={Zhang, Xiangyu},
      author={Ren, Shaoqing},
      author={Sun, Jian},
       title={{Identity Mappings in Deep Residual Networks}},
        date={201603},
      series={36th International Conference on Machine Learning},
      volume={cs.CV},
       pages={630 645},
}

\bib{KidgerLyons2019}{article}{
      author={Kidger, Patrick},
      author={Lyons, Terry},
       title={{Universal Approximation with Deep Narrow Networks}},
        date={2019},
     journal={arXiv},
      eprint={1905.08539},
}

\bib{kidger2020neural}{article}{
      author={Kidger, Patrick},
      author={Morrill, James},
      author={Foster, James},
      author={Lyons, Terry},
       title={Neural controlled differential equations for irregular time series},
        date={2020},
     journal={Advances in Neural Information Processing Systems},
      volume={33},
       pages={6696\ndash 6707},
}

\bib{Kobyzev2020NormalizingFA}{article}{
      author={Kobyzev, Ivan},
      author={Prince, Simon},
      author={Brubaker, Marcus~A.},
       title={Normalizing flows: An introduction and review of current methods},
        date={2020},
     journal={IEEE Transactions on Pattern Analysis and Machine Intelligence},
      volume={43},
       pages={3964\ndash 3979},
         url={https://api.semanticscholar.org/CorpusID:208910764},
}

\bib{Li:2017wr}{article}{
      author={Li, Qianxiao},
      author={Chen, Long},
      author={Tai, Cheng},
      author={Weinan, E},
       title={Maximum principle based algorithms for deep learning},
        date={2018},
     journal={Journal of Machine Learning Research},
      volume={18},
      number={165},
       pages={1\ndash 29},
}

\bib{Lin:2020wv}{article}{
      author={Lin, Alex~Tong},
      author={Fung, Samy~Wu},
      author={Li, Wuchen},
      author={Nurbekyan, Levon},
      author={Osher, Stanley~J},
       title={Alternating the population and control neural networks to solve high-dimensional stochastic mean-field games},
        date={2021},
     journal={Proceedings of the National Academy of Sciences},
      volume={118},
      number={31},
       pages={e2024713118},
}

\bib{LasryLions2007}{article}{
      author={Lasry, Jean-Michel},
      author={Lions, Pierre-Louis},
       title={Mean field games},
        date={200703},
     journal={Japanese Journal of Mathematics},
      volume={2},
       pages={229\ndash 260},
}

\bib{Li:2019wr}{article}{
      author={Li, Qianxiao},
      author={Lin, Ting},
      author={Shen, Zuowei},
       title={Deep learning via dynamical systems: An approximation perspective},
        date={2022},
     journal={Journal of the European Mathematical Society},
      volume={25},
      number={5},
       pages={1671\ndash 1709},
}

\bib{massaroli2020dissecting}{inproceedings}{
      author={Massaroli, Stefano},
      author={Poli, Michael},
      author={Park, Jinkyoo},
      author={Yamashita, Atsushi},
      author={Asama, Hajime},
       title={Dissecting neural odes},
        date={2020},
   booktitle={Advances in neural information processing systems},
      volume={33},
       pages={3952\ndash 3963},
}

\bib{OnkenEtAl2020OTFlow}{inproceedings}{
      author={Onken, Derek},
      author={Fung, Samy~Wu},
      author={Li, Xingjian},
      author={Ruthotto, Lars},
       title={{OT-Flow}: Fast and accurate continuous normalizing flows via optimal transport},
        date={2021},
   booktitle={Proceedings of the {AAAI} {C}onference on {A}rtificial {I}ntelligence},
      volume={35},
       pages={9223\ndash 9232},
}

\bib{onken2020discretize}{article}{
      author={Onken, Derek},
      author={Ruthotto, Lars},
       title={Discretize-optimize vs. optimize-discretize for time-series regression and continuous normalizing flows},
        date={2020},
     journal={arXiv preprint arXiv:2005.13420},
}

\bib{Ruthotto2018DeepNN}{article}{
      author={Ruthotto, Lars},
      author={Haber, Eldad},
       title={Deep neural networks motivated by partial differential equations},
        date={2018},
     journal={Journal of Mathematical Imaging and Vision},
      volume={62},
       pages={352 \ndash  364},
         url={https://api.semanticscholar.org/CorpusID:4788755},
}

\bib{RuthottoHaber2021}{article}{
      author={Ruthotto, Lars},
      author={Haber, Eldad},
       title={An introduction to deep generative modeling},
        date={2021},
     journal={GAMM-Mitteilungen},
      volume={44},
      number={2},
       pages={e202100008},
}

\bib{GonzalezGarcia:2004wp}{article}{
      author={Rico-Martinez, Ramiro},
      author={Krischer, K},
      author={Kevrekidis, IG},
      author={Kube, MC},
      author={Hudson, JL},
       title={Discrete-vs. continuous-time nonlinear signal processing of cu electrodissolution data},
        date={1992},
     journal={Chemical Engineering Communications},
      volume={118},
      number={1},
       pages={25\ndash 48},
}

\bib{RuthottoEtAl2020MFG}{article}{
      author={Ruthotto, Lars},
      author={Osher, S~J},
      author={Li, Wuchen},
      author={Nurbekyan, Levon},
      author={Fung, Samy~Wu},
       title={{A machine learning framework for solving high-dimensional mean field game and mean field control problems}},
        date={2020},
     journal={Proceedings of the National Academy of Sciences},
      volume={117},
      number={17},
       pages={9183 \ndash  9193},
}

\bib{Sander2022}{article}{
      author={Sander, Michael~E},
      author={Ablin, Pierre},
      author={Peyré, Gabriel},
       title={{Do Residual Neural Networks discretize Neural Ordinary Differential Equations?}},
        date={2022},
     journal={arXiv},
      eprint={2205.14612},
}

\bib{WangEtAl2018}{article}{
      author={Wang, Bao},
      author={Yuan, Binjie},
      author={Shi, Zuoqiang},
      author={Osher, Stanley~J.},
       title={Enresnet: Resnets ensemble via the feynman--kac formalism for adversarial defense and beyond},
        date={2020},
     journal={SIAM Journal on Mathematics of Data Science},
      volume={2},
      number={3},
       pages={559\ndash 582},
      eprint={https://doi.org/10.1137/19M1265302},
         url={https://doi.org/10.1137/19M1265302},
}

\bib{Yang:2019tj}{article}{
      author={Yang, Liu},
      author={Karniadakis, George~Em},
       title={Potential flow generator with {$L_2$} optimal transport regularity for generative models},
        date={2020},
     journal={IEEE Transactions on Neural Networks and Learning Systems},
      volume={33},
      number={2},
       pages={528\ndash 538},
}

\bib{ZhangEtAl2023}{article}{
      author={Zhang, Benjamin~J},
      author={Katsoulakis, Markos~A},
       title={{A mean-field games laboratory for generative modeling}},
        date={2023},
     journal={arXiv},
      eprint={2304.13534},
}

\end{biblist}
\end{bibdiv}
